\documentclass[letterpaper, 10 pt, conference]{ieeeconf}  

\IEEEoverridecommandlockouts                              

\usepackage{graphicx}
\usepackage{amsmath}
\usepackage{amssymb}
\usepackage{booktabs}
\usepackage{multirow}
\usepackage[table]{xcolor}
\usepackage{marvosym}
\usepackage[hidelinks]{hyperref}

\title{\LARGE \bf
CAR-VLA: Complexity-Aware and Risk-Adaptive Reasoning for Autonomous Driving
}
\author{Xiaolei Chen$^{1}$, Zhuolin He$^{1,2,\dagger}$, Yuxuan Liang$^{1}$, Xu Li$^{1}$, Haotian Chen$^{1}$, Fan Shi$^{3}$, Mengyang Zhao$^{1}$,\\ Wenjuan Meng$^{1}$, Zisheng Chen$^{4}$, Zhihao Zhu$^{2}$, Zhounan Jin$^{5}$, Hengli Wang$^{5}$,\\ Qingfan Wang$^{2}$, Jiamei Liang$^{2}$, Bin Li$^{1}$, Xiangyang Xue\textsuperscript{1,\Letter}
\thanks{$^{1}$Fudan University, $^{2}$Yinwang Intelligent Technology Co., Ltd, $^{3}$Fuzhou University, $^{4}$Sun Yat-Sen University, $^{5}$Huawei Technology}%
\thanks{Email: chenxl23@m.fudan.edu.cn}%
\thanks{$\dagger$: Projector Leader.  \qquad \qquad \Letter: Corresponding Author.}%
}

\begin{document}

\maketitle
\thispagestyle{empty}
\pagestyle{empty}

\begin{abstract}

Existing adaptive reasoning methods for driving Vision-Language-Action (VLA) models primarily focus on whether to reason, overlooking how reasoning should differ across driving situations. Our key insight is that while scene complexity informs reasoning depth, dynamic risk is equally critical for deciding how to reason in time-critical situations. We therefore propose CAR-VLA, a unified driving VLA model that jointly considers scene complexity and dynamic risk to guide reasoning depth, urgency, and focus. CAR-VLA maps four complexity--risk categories to three reasoning modes: \textit{Fast Intuition} for direct trajectory generation in simple low-risk scenes, \textit{Slow Thinking} for deliberate reasoning in complex low-risk scenes, and \textit{Reflex Response} for compact, hazard-focused reasoning in high-risk scenes regardless of complexity. Rather than merely shortening deliberation, Reflex Response centers reasoning on the most critical hazard and the immediate safe response. We train CAR-VLA through progressive supervised learning that links scene assessment, reasoning-mode selection, and trajectory generation, followed by reasoning-augmented reinforcement learning to improve driving quality and reasoning behavior. Experiments on NAVSIM v1(91.1 PDMS), NAVSIM v2(90.3 EPDMS), and Navhard(35.0 EPDMS) demonstrate competitive driving performance. Qualitative comparisons on navtest and in-house high-risk scenarios further illustrate risk-aware reasoning and hazard-responsive trajectory generation. The code for this paper will be released publicly at: \url{https://github.com/chenxl124578/CAR-VLA.git}.
\end{abstract}

\section{INTRODUCTION}
\label{sec:introduction}
Vision-Language-Action (VLA) models are increasingly used in embodied agent intelligence to connect visual perception, language understanding, and action generation within a unified framework~\cite{zhou2026opendrivevla,wang2026linkvla,huang2023vlmaps}. Recent driving VLAs further incorporate explicit reasoning to improve planning quality and decision making~\cite{bassole2026hybriddrivevla,yuan2026autodriver2}. As autonomous driving moves toward real-world deployment, safety, reliability, and timely decision making remain central requirements~\cite{feng2026breaking}. This creates a distinct challenge for reasoning-based driving models: while some situations benefit from deliberate analysis, time-critical hazards require rapid and focused responses. Therefore, adaptive reasoning should not only determine whether more reasoning is needed, but also what form of reasoning best matches the current driving situation.

\begin{figure}
    \centering
    \includegraphics[width=0.90\linewidth]{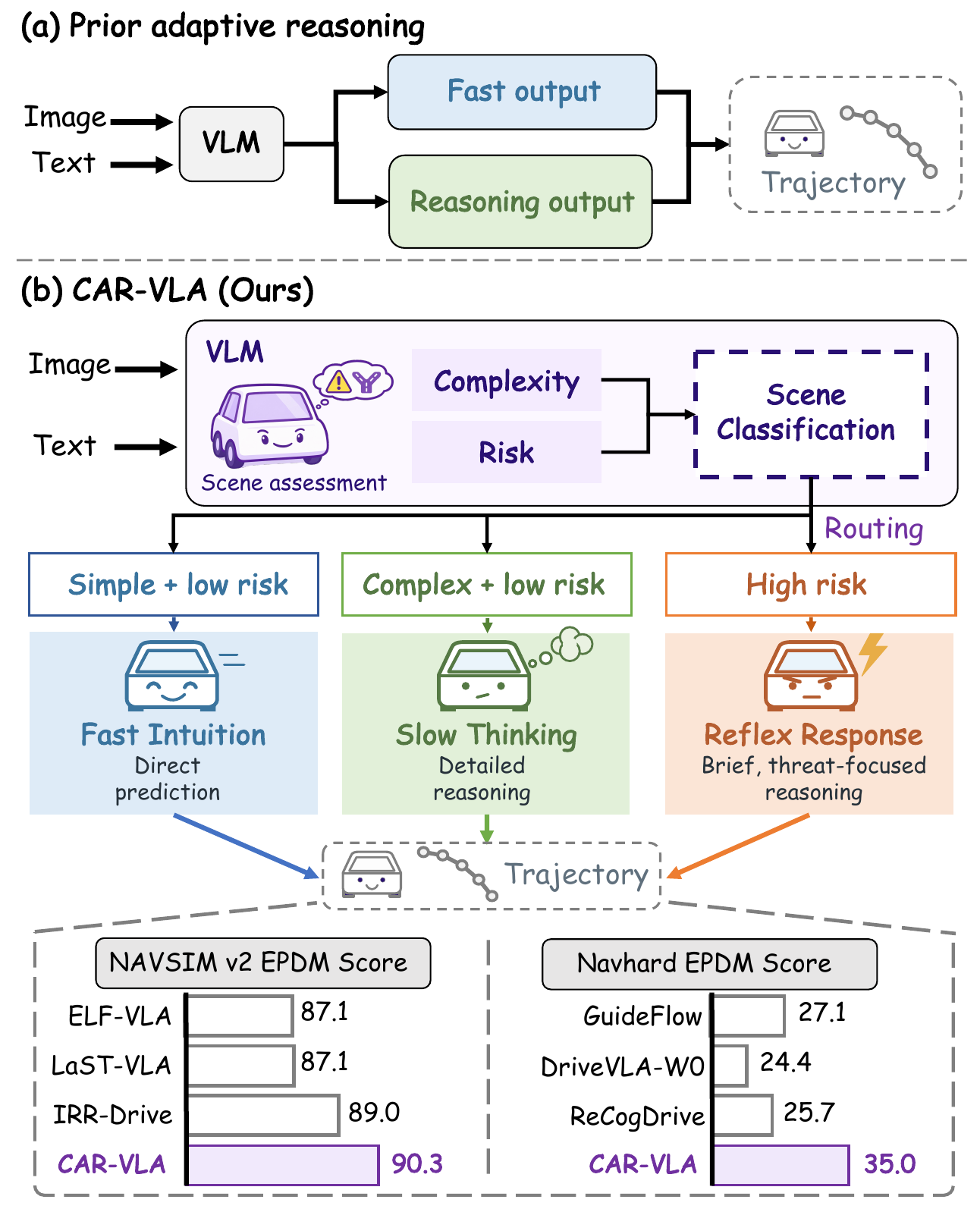}
    \caption{{\textbf{Comparison of Reasoning-based VLA Paradigms.}} (a) {\textbf{Prior adaptive reasoning}} methods adaptively choose whether to perform reasoning, without recognizing that different scenes require different reasoning styles. (b) {\textbf{Our CAR-VLA}} analyzes scene complexity and risk, and routes different scene categories to different reasoning modes. These choices represent different reasoning focuses and depths, satisfying the reasoning requirements of diverse scenarios.  The results below demonstrate CAR-VLA’s superior performance on the NAVSIM leaderboard.
    }
    \label{introfig}
\end{figure}
Existing reasoning-based driving VLA primarily addresses this challenge through adaptive selection of reasoning, as shown in Fig.\ref{introfig}(a). AutoVLA~\cite{zhou2025autovla} learns fast trajectory-only prediction and slow chain-of-thought reasoning, while AdaThinkDrive~\cite{luo2025adathinkdrive} encourages the model to invoke slow reasoning only when it improves trajectory quality. IRR-Drive~\cite{chen2026irrdrive} similarly switches between direct prediction and reflective refinement according to scene complexity. Although these methods all consider the use of reasoning to optimize actions, they primarily place adaptation in a single dimension: whether the scenario requires additional reasoning to optimize the trajectory. This overlooks an important distinction in driving: a complex junction may require more analysis even when there is no immediate safety threat, whereas a sudden cut-in on a simple road may require less deliberation but a faster and more focused response. 

{This distinction leads to our key insight: {\textbf{while scene complexity is an important factor in determining reasoning depth, dynamic risk provides an equally critical dimension for deciding how to reason in time-critical driving situations}}. Scene complexity reflects the difficulty of understanding and planning in a scene, arising mainly from road topology, traffic rules, and spatial layout. However, complexity alone does not capture the urgency of interactions with surrounding objects. Dynamic risk complements this perspective by characterizing the imminence of potential conflicts, thereby informing how quickly and with what focus the model should respond. In this sense, complexity primarily shapes the need for deliberation, while risk shapes the urgency and focus of reasoning. Together, these factors motivate different reasoning regimes: simple low-risk scenes can be handled with direct planning, complex but low-risk scenes benefit from deliberate reasoning, and high-risk scenes call for compact, hazard-focused reasoning. {\textbf{Our central contribution is therefore to extend adaptive driving reasoning beyond complexity-based depth adaptation by explicitly incorporating dynamic risk, allowing both dimensions to jointly guide reasoning depth, urgency, and focus.}}}

Based on this formulation, we propose \textbf{CAR-VLA}, a unified driving VLA that adapts its reasoning according to scene complexity and dynamic risk, as shown in Fig.~\ref{introfig}(b). CAR-VLA maps four complexity--risk categories to three reasoning modes: \textit{Fast Intuition} for direct trajectory generation in simple low-risk scenes, \textit{Slow Thinking} for detailed reasoning in complex low-risk scenes, and \textit{Reflex Response} for brief, threat-focused reasoning in high-risk scenes, regardless of scene complexity. Unlike simply shortening deliberative reasoning, Reflex Response reorganizes the reasoning process around the most critical hazard and the immediate safe response. We train CAR-VLA through progressive supervised learning to link scene assessment, reasoning-mode selection, and trajectory generation, followed by reasoning-augmented Reinforcement Learning (RL) to improve driving quality and reasoning behavior. Our contributions are as follows:
\begin{itemize}
\item We formulate adaptive driving reasoning around two complementary factors, scene complexity and dynamic risk, which capture different demands on deliberation and response urgency.
\item We propose CAR-VLA, a unified VLA framework that adapts among Fast Intuition, Slow Thinking, and Reflex Response which includes a dedicated hazard-focused reasoning mode for high-risk situations.
\item We develop a progressive learning strategy that links scene assessment, reasoning-mode selection, and trajectory generation, and further improves driving performance through reasoning-augmented RL.
\item Extensive experiments demonstrating competitive performance on NAVSIM v1 (91.1 PDMS), NAVSIM v2 (90.3 EPDMS) and Navhard (35.0 EPDMS). Qualitative comparisons on Navtest and in-house high risk scenario further illustrate CAR-VLA’s risk-aware reasoning and its generation of trajectories that respond to identified hazards.
\end{itemize}
\section{RELATED WORK}
\label{sec:related-work}


\subsection{Vision-Language-Action Driving}
VLA models connect scene understanding with action generation through multimodal modeling and language-action alignment~\cite{hwang2025emma,renz2025simlingo,jiang2026senna,fu2025orion}. OpenDriveVLA~\cite{zhou2026opendrivevla} introduces hierarchical vision-language alignment and structured interaction modeling, while LinkVLA~\cite{wang2026linkvla} uses shared discrete representations and coarse-to-fine action decoding. To enhance reasoning and planning, AutoDrive-R$^2$~\cite{yuan2026autodriver2} combines chain-of-thought reasoning, self-reflection, and reinforcement learning, while HybridDriveVLA~\cite{bassole2026hybriddrivevla} incorporates visual chain-of-thought reasoning and Tree-of-Thought-inspired waypoint evaluation. 
These studies show a clear trend toward using VLA models as unified driving systems that connect semantic understanding with trajectory generation.


\subsection{Adaptive Reasoning for Autonomous Driving}
Recent studies have begun to question whether explicit reasoning should be applied uniformly across all driving scenes. AutoVLA~\cite{zhou2025autovla} introduces fast and slow thinking modes, using trajectory-only prediction for straightforward scenes and chain-of-thought reasoning when additional reasoning is needed. Counterfactual VLA~\cite{peng2026cfvla} further adopts adaptive thinking by selectively activating counterfactual self-reflection in challenging scenarios. FutureSightDrive~\cite{zeng2026futuresightdrive} introduces visual spatio-temporal CoT that captures future scene evolution and spatial relationships to guide trajectory planning. Meanwhile, Latent-CoT-Drive~\cite{tan2026latentcot} replaces text-based reasoning with action-aligned latent reasoning to improve inference efficiency, while NoRD~\cite{rawal2026nord} shows that competitive driving performance can also be achieved without explicit reasoning, highlighting that reasoning is not always necessary. However, existing adaptive approaches mainly adjust the amount or representation of reasoning according to scene difficulty. They do not explicitly separate \emph{scene complexity}, which determines the need for deliberation, from \emph{dynamic risk}, which determines the urgency of action. Our work focuses on this distinction and uses both factors to determine the reasoning process for different driving situations.

\section{METHOD}
\label{sec:method}

\begin{figure*}[t]
    \centering
    \includegraphics[width=0.95\linewidth]{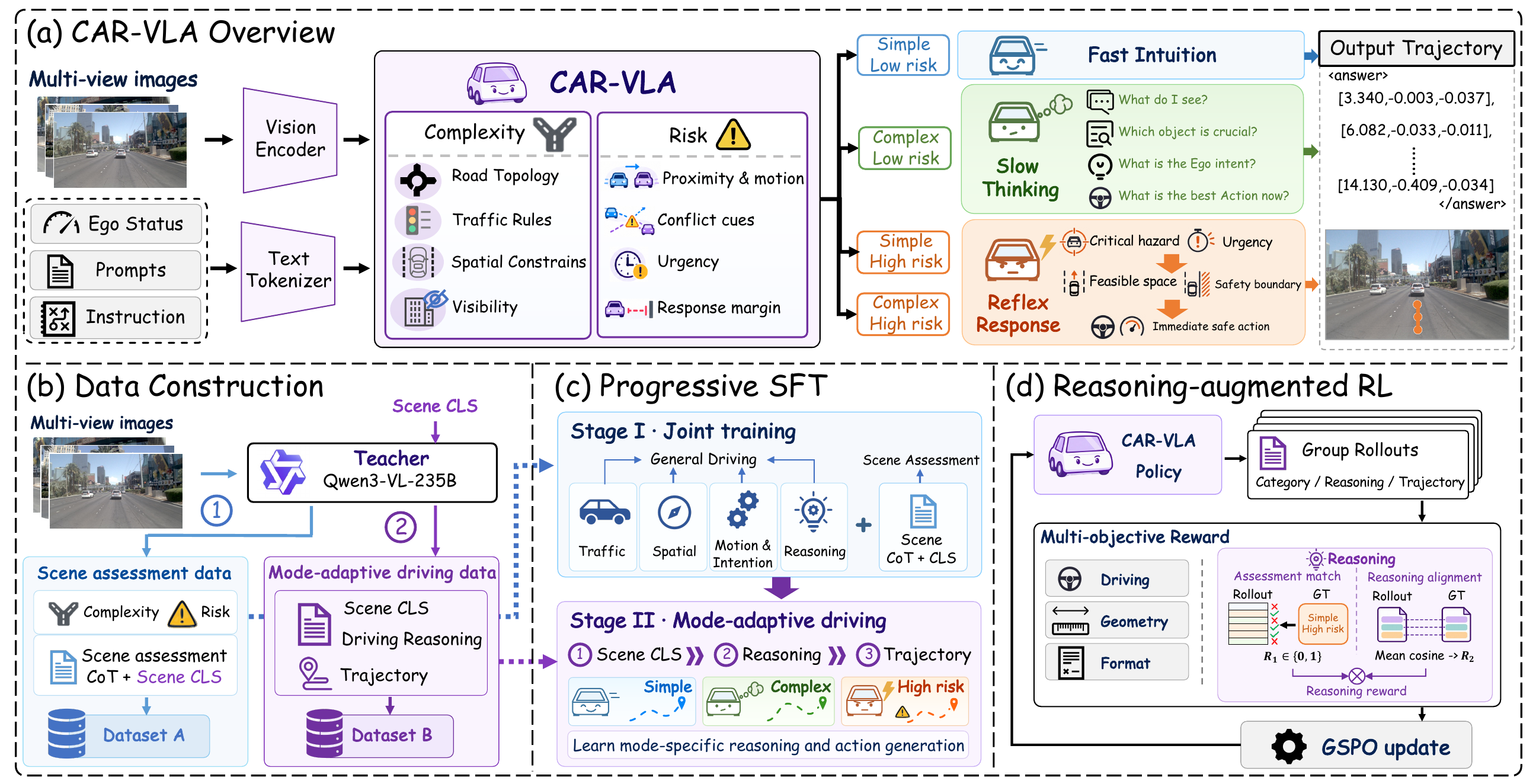}
    \caption{
\textbf{Overview of CAR-VLA.}
\textbf{(a)} CAR-VLA separately assesses scene complexity and risk, mapping four scene categories to three driving modes: \emph{Fast Intuition} for simple, low-risk scenes, \emph{Slow Thinking} for complex, low-risk scenes, and \emph{Reflex Response} for high-risk scenes regardless of complexity.
The selected mode guides subsequent trajectory generation.
\textbf{(b)} Teacher-assisted data construction produces two complementary datasets: scene assessment data containing assessment reasoning and scene labels, and mode-adaptive driving data containing scene labels, driving reasoning, and trajectories.
\textbf{(c)} Progressive supervised fine-tuning jointly learns general driving knowledge and scene assessment in Stage~I, then learns scene classification, mode-specific reasoning, and trajectory generation in Stage~II.
\textbf{(d)} Reasoning-augmented reinforcement learning updates the policy through GSPO using driving, geometry, format, and reasoning rewards.
The reasoning reward combines scene-assessment correctness with alignment to reference reasoning.
}
    \label{fig:method_overview}
\end{figure*}

CAR-VLA dynamically adapts its driving reasoning strategy according to scene complexity and dynamic risk. The unified VLA model maps four complexity--risk scene categories to three reasoning modes: \textit{Fast Intuition} for simple low-risk scenes, \textit{Slow Thinking} for complex low-risk scenes, and \textit{Reflex Response} for high-risk scenes, as shown in Fig.~\ref{fig:method_overview}(a). We next formulate this adaptive reasoning framework and describe its data construction and training.

\subsection{Problem Formulation}
\label{sec:problem_formulation}

Given camera observations $\mathcal{I}_t$, ego state
$\mathbf{s}_t$, and navigation instruction $u_t$, we define
the driving query as
$q_t=(\mathcal{I}_t,\mathbf{s}_t,u_t)$.
The planning task is to predict a future ego trajectory
\begin{equation}
\hat{\boldsymbol{\tau}}_t
=
\bigl(
(\hat{x}_{t,k},\hat{y}_{t,k},\hat{\psi}_{t,k})
\bigr)_{k=1}^{K},
\label{eq:trajectory}
\end{equation}
where each pose specifies the ego position and heading
at time $t+k\Delta t$. We use $K=10$ and
$\Delta t=0.5\,\mathrm{s}$, yielding a five-second
planning horizon.

We characterize each scene by a category
$z_t=(C_t,D_t)\in\{0,1\}^{2}$, where $C_t=0/1$
denotes simple/complex scenes and $D_t=0/1$ denotes
low/high dynamic risk.
Complexity is assessed from static or quasi-static
context, whereas risk concerns short-term interaction
threats and the available margin for a safe response.
The four complexity--risk categories are mapped to
three reasoning modes:
\begin{equation}
f(z_t)=
\begin{cases}
\text{Fast Intuition},
    & z_t=(0,0),\\
\text{Slow Thinking},
    & z_t=(1,0),\\
\text{Reflex Response},
    & z_t\in\{(0,1),(1,1)\}.
\end{cases}
\label{eq:mode_mapping}
\end{equation}
Thus, the two high-risk categories remain distinct
scene labels but share the same reasoning mode.

CAR-VLA uses a single autoregressive policy to generate
the scene category, mode-specific driving reasoning,
and trajectory in sequence:
\begin{equation}
y_t=
\bigl(
\hat{z}_t,c_t,\hat{\boldsymbol{\tau}}_t
\bigr)
\sim \pi_{\theta}(\cdot\mid q_t),
\label{eq:policy_output}
\end{equation}
where $\hat{z}_t$ is the predicted scene category and
$c_t$ denotes the driving reasoning associated with
$f(\hat{z}_t)$.
For Fast Intuition, $c_t=\varnothing$, so the model
generates the trajectory directly after the scene
category.

\subsection{Complexity--Risk Guided Data Construction}
\label{sec:data_curation}

Complexity--risk-guided supervision is constructed from 103k \textit{navtrain} scenes~\cite{caesar2021nuplan,dauner2024navsim}, as illustrated in Fig.~\ref{fig:method_overview}(b). Each multi-view driving scene is annotated by a Qwen3-VL-235B teacher~\cite{bai2025qwen3vl} for scene assessment and mode-specific driving reasoning.

The teacher assesses \textit{scene complexity} and \textit{dynamic risk} separately. Scene complexity captures road topology, traffic rules, spatial constraints, and visibility, while dynamic risk reflects relative motion, potential conflicts, and response urgency. Dimension-specific evidence is required for each assessment, so environmental complexity alone does not indicate high risk. The annotations provide \textit{scene assessment reasoning} and the corresponding scene category (represented by Scene assessment CoT and SceneCLS respectively in Fig.~\ref{fig:method_overview}(b)).

The scene category determines the driving reasoning according to Eq.~\eqref{eq:mode_mapping}. Simple low-risk scenes use \textit{Fast Intuition} without explicit reasoning, while complex low-risk scenes use the \textit{Slow Thinking} template from AutoVLA~\cite{zhou2025autovla}. For high-risk scenes, we introduce a compact \textit{Reflex Response} template with five components: \textit{Hazard}, \textit{Urgency}, \textit{Feasibility}, \textit{Constraint}, and \textit{Action}. It focuses on the most critical hazard and its urgency, feasible maneuvers and safety boundary, and the immediate response. Unlike extended deliberation, \textit{Reflex Response} prioritizes the critical conflict and available response options.

The final driving target contains the scene category, mode-specific driving reasoning, and ground-truth trajectory, without the preceding scene assessment reasoning.

\subsection{Progressive Supervised Fine-Tuning}
\label{sec:sft}

As shown in Fig.~\ref{fig:method_overview}(c), CAR-VLA is fine-tuned in two stages: general driving understanding and scene assessment are jointly learned first, followed by mode-adaptive driving fine-tuning.

CAR-VLA is jointly fine-tuned on general driving data from ReCogDrive~\cite{li2025recogdrive} and the scene assessment annotations described above. The ReCogDrive data covers traffic understanding, spatial relations, motion and intention understanding, and driving reasoning. Scene assessment annotations supervise the reasoning and classification of scene complexity and dynamic risk. This joint training equips the model with general driving knowledge and complexity--risk assessment capability.

\begin{table*}[!t]
    \centering
    \caption{Performance Comparison on Navtest split in NAVSIM v1 and v2 using Closed-Loop Metrics.}
    \label{tab:navsim_main}
    \label{tab:navsim_v2}
    \label{tab:navsim_v1}
    \small
    \newcommand{\navsimsep}[1]{\multicolumn{1}{@{\hspace{2pt}\vrule width 0.4pt\hspace{2pt}}c}{#1}}
    \newcommand{\navsimlast}[1]{\multicolumn{1}{@{\hspace{2pt}\vrule width 0.4pt\hspace{2pt}}c@{}}{#1}}
    \begin{tabular}{@{}l*{16}{@{\hspace{2pt}}c}@{}}
        \toprule
        \multirow{2}{*}{Method} & \multicolumn{6}{c}{NAVSIM v1} & \multicolumn{10}{c}{NAVSIM v2} \\
        \cmidrule(lr){2-7}\cmidrule(lr){8-17}
        & \navsimsep{NC$\uparrow$} & DAC$\uparrow$ & EP$\uparrow$ & TTC$\uparrow$ & C.$\uparrow$ & \navsimsep{PDMS$\uparrow$}
        & \navsimsep{NC$\uparrow$} & DAC$\uparrow$ & DDC$\uparrow$ & TLC$\uparrow$ & EP$\uparrow$ & TTC$\uparrow$ & LK$\uparrow$ & HC$\uparrow$ & EC$\uparrow$ & \navsimlast{EPDMS$\uparrow$} \\
        \midrule
        \rowcolor{gray!12}\multicolumn{17}{c}{\emph{Traditional End-to-End Methods}} \\
        HydraMDP++~\cite{li2025hydramdppp}
            & \navsimsep{97.6} & 96.0 & 80.4 & 93.1 & 100.0 & \navsimsep{86.6}
            & \navsimsep{97.2} & 97.5 & 99.4 & 99.6 & 83.1 & 96.5 & 94.4 & 98.2 & 70.9 & \navsimlast{81.4} \\
        DriveSuprim~\cite{yao2025drivesuprim}
            & \navsimsep{97.8} & 97.3 & 86.7 & 93.6 & 100.0 & \navsimsep{89.9}
            & \navsimsep{97.5} & 96.5 & 99.4 & 99.6 & 88.4 & 96.6 & 95.5 & 98.3 & 77.0 & \navsimlast{83.1} \\
        DiffusionDrive~\cite{liao2025diffusiondrive}
            & \navsimsep{98.2} & 96.2 & 82.2 & 94.7 & 100.0 & \navsimsep{88.1}
            & \navsimsep{98.2} & 95.9 & 99.4 & 99.8 & 87.5 & 97.3 & 96.8 & 98.3 & 87.7 & \navsimlast{84.5} \\
        ResAD~\cite{zheng2025resad}
            & \navsimsep{98.0} & 97.5 & 83.3 & 94.1 & 100.0 & \navsimsep{88.8}
            & \navsimsep{97.8} & 97.2 & 99.5 & 99.8 & 88.2 & 96.9 & 97.0 & 98.4 & 88.2 & \navsimlast{85.5} \\
        \midrule
        \rowcolor{gray!12}\multicolumn{17}{c}{\emph{VLA Methods Without Explicit Reasoning}} \\
        ReCogDrive~\cite{li2025recogdrive}
            & \navsimsep{98.2} & 97.5 & 83.5 & 95.2 & 99.9 & \navsimsep{89.6}
            & \navsimsep{98.3} & 95.2 & 99.5 & 99.8 & 87.1 & 97.5 & 96.6 & 98.3 & 86.5 & \navsimlast{83.6} \\
        DriveVLA-W0~\cite{li2025drivevlaw0}
            & \navsimsep{98.7} & 99.1 & 83.3 & 95.3 & 99.3 & \navsimsep{90.3}
            & \navsimsep{98.5} & 99.1 & 98.0 & 99.7 & 86.4 & 98.1 & 93.2 & 97.9 & 58.9 & \navsimlast{86.1} \\
        DriveFine~\cite{dang2026drivefine}
            & \navsimsep{98.6} & 97.9 & 85.5 & 95.2 & 99.9 & \navsimsep{90.7}
            & \navsimsep{98.7} & 97.3 & 98.8 & 99.8 & 88.2 & 97.8 & 97.7 & 98.4 & 84.7 & \navsimlast{87.1} \\
        SGDrive~\cite{li2026sgdrive}
            & \navsimsep{98.6} & 97.8 & 85.8 & 96.2 & 100.0 & \navsimsep{\underline{91.1}}
            & \navsimsep{98.6} & 94.3 & 99.5 & 99.9 & 86.0 & 97.9 & 96.1 & 98.3 & 85.9 & \navsimlast{86.2} \\
        \midrule
        \rowcolor{gray!12}\multicolumn{17}{c}{\emph{VLA Methods With Explicit Reasoning}} \\
        AutoVLA$^{\dagger}$~\cite{zhou2025autovla}
            & \navsimsep{98.4} & 95.6 & 81.9 & 98.0 & 99.9 & \navsimsep{89.1}
            & \navsimsep{98.9} & 94.8 & 99.0 & 99.9 & 86.9 & 98.1 & 97.7 & 98.0 & 84.3 & \navsimlast{87.6} \\
        ELF-VLA~\cite{luo2026elfvla}
            & \navsimsep{98.9} & 98.1 & 85.3 & 96.0 & 100.0 & \navsimsep{91.0}
            & \navsimsep{98.9} & 98.1 & 99.4 & 99.8 & 88.5 & 98.4 & 96.9 & 98.3 & 87.2 & \navsimlast{87.1} \\
        LaST-VLA~\cite{luo2026lastvla}
            & \navsimsep{98.7} & 97.9 & 86.8 & 95.6 & 100.0 & \navsimsep{\textbf{91.3}}
            & \navsimsep{98.7} & 97.9 & 99.2 & 99.7 & 90.3 & 98.2 & 96.6 & 98.3 & 86.3 & \navsimlast{87.1} \\
        IRR-Drive~\cite{chen2026irrdrive}
            & \navsimsep{98.0} & 98.3 & 88.5 & 93.7 & 100.0 & \navsimsep{\textbf{91.3}}
            & \navsimsep{97.0} & 98.3 & 98.9 & 99.5 & 92.3 & 96.8 & 95.8 & 97.6 & 82.2 & \navsimlast{\underline{89.0}} \\
        AdaThinkDrive~\cite{luo2025adathinkdrive}
            & \navsimsep{98.4} & 97.8 & 84.4 & 95.2 & 100.0 & \navsimsep{90.3}
            & \navsimsep{--} & -- & -- & -- & -- & -- & -- & -- & -- & \navsimlast{--} \\
        \midrule
        \textbf{CAR-VLA-SFT (Ours)}
            & \navsimsep{98.4} & 96.6 & 81.3 & 95.3 & 100.0 & \navsimsep{88.1}
            & \navsimsep{98.6} & 96.6 & 99.5 & 99.9 & 86.3 & 98.0 & 97.4 & 98.3 & 86.0 & \navsimlast{88.5} \\
        \textbf{CAR-VLA-RL (Ours)}
            & \navsimsep{98.4} & 98.1 & 86.7 & 95.0 & 100.0 & \navsimsep{\underline{91.1}}
            & \navsimsep{98.5} & 98.1 & 99.4 & 99.9 & 89.0 & 97.9 & 97.4 & 98.3 & 85.8 & \navsimlast{\textbf{90.3}} \\
        \bottomrule
    \end{tabular}
    \par\vspace{2pt}
    \noindent
    \begin{minipage}{0.95\linewidth}
        \footnotesize
        \raggedright
        C. denotes comfort. $^{\dagger}$ means NAVSIM v2 results reproduced using the official checkpoint. ${\uparrow}$ indicates higher is better. The best and second best results are \textbf{bold} and \underline{underlined}, respectively.
    \end{minipage}
\end{table*}

The resulting model is further fine-tuned with the mode-adaptive driving annotations. Each target contains the scene category, mode-specific driving reasoning, and ground-truth trajectory. For \textit{Fast Intuition}, the reasoning component is omitted and the trajectory is generated directly after the scene category. All three reasoning modes are learned within the same autoregressive policy.

Both stages minimize the autoregressive
negative log-likelihood of the target response:
\begin{equation}
\mathcal{L}_{\mathrm{SFT}}(\theta;\mathcal{D})
=
-\mathbb{E}_{(q,y)\sim\mathcal{D}}
\left[
\sum_{j=1}^{|y|}
\log \pi_{\theta}(y_j \mid q,y_{<j})
\right],
\label{eq:sft_loss}
\end{equation}
where $y$ denotes the target response and $\mathcal{D}$ denotes the training data used in each step. The resulting SFT policy initializes the subsequent reinforcement-learning stage.

\subsection{Reasoning-Augmented Reinforcement Learning}
\label{sec:rl}

The SFT policy is further optimized through GSPO~\cite{gspo}, using feedback on driving performance and reasoning quality. As shown in Fig.~\ref{fig:method_overview}(d), the total reward combines driving, geometry, and reasoning rewards under a binary format gate:
\begin{equation}
\begin{aligned}
R(q,y)
&= R_{\mathrm{format}}\Bigl(
\lambda_{\mathrm{d}}R_{\mathrm{drive}}
+ \lambda_{\mathrm{g}}R_{\mathrm{geometry}}
+ \lambda_{\mathrm{r}}R_{\mathrm{reason}}
\Bigr),
\end{aligned}
\label{eq:total_reward}
\end{equation}
where $\lambda_{\mathrm{d}}$, $\lambda_{\mathrm{g}}$, and $\lambda_{\mathrm{r}}$ are nonnegative reward weights.

The driving reward $R_{\mathrm{drive}}\in[0,1]$ is the PDMS of the predicted trajectory evaluated in NAVSIM~\cite{dauner2024navsim}. The geometry reward $R_{\mathrm{geometry}}$ is the concentric OBB-FDE reward from IRR-Drive~\cite{chen2026irrdrive}. 
The format reward $R_{\mathrm{format}}\in\{0,1\}$ checks the mode-specific output structure, including the use of \texttt{<type>}, \texttt{<think>}, and \texttt{<answer>} tags. The trajectory must parse into exactly ten numerical poses. Responses that satisfy these requirements receive $R_{\mathrm{format}}=1$. Any violation sets the total reward to zero, without evaluating the remaining reward terms.

We introduce the reasoning reward combines scene-category correctness with section-wise reasoning alignment. Alignment is evaluated only when the predicted scene category matches the annotation:
\begin{equation}
\begin{aligned}
R_{\mathrm{cls}}
&= \mathbf{1}[\hat{z}=z^{*}],
R_{\mathrm{reason}}
=
\begin{cases}
R_{\mathrm{align}}, & R_{\mathrm{cls}}=1, \\
0,                 & R_{\mathrm{cls}}=0,
\end{cases}
\end{aligned}
\label{eq:reasoning_reward}
\end{equation}
where $\hat{z}$ and $z^{*}$ denote the predicted and annotated scene categories, respectively.

For correctly classified \textit{Slow Thinking} and \textit{Reflex Response} samples, the generated reasoning is compared with the reference annotation under matching section headings:
\begin{equation}
R_{\mathrm{align}}
=
\left[
\frac{1}{|\mathcal{H}|}
\sum_{h\in\mathcal{H}}
\cos\!\left(E(c_h),E(c_h^{*})\right)
\right]_{+},
\label{eq:reasoning_alignment}
\end{equation}
where $\mathcal{H}$ is the set of section headings in the reference reasoning template. The texts $c_h$ and $c_h^{*}$ contain the generated and reference reasoning under heading $h$, and $E$ is a sentence embedding encoder. The operator $[a]_{+}=\max(0,a)$ clips the mean cosine similarity at zero, giving $R_{\mathrm{align}}\in[0,1]$. Each reasoning section is compared with its corresponding reference rather than scoring the entire trace as a single text.

\textit{Fast Intuition} requires no explicit driving reasoning, so $R_{\mathrm{align}}$ is set to $1$ for correctly classified samples. Its reasoning reward therefore reduces to scene-category correctness. Any incorrect category receives zero reasoning reward, including confusion between the two high-risk categories that share \textit{Reflex Response}. Category errors do not affect the driving or geometry rewards for format-valid responses.

During policy optimization, a group of responses $\{y_i\}_{i=1}^{G}$ is sampled from the rollout policy $\pi{\theta_{\mathrm{old}}}(\cdot\mid q)$ for each query $q$ and scored using Eq.~\eqref{eq:total_reward}. The rewards are normalized within each group to obtain relative advantages. The policy is then optimized with GSPO, using length-normalized sequence likelihood ratios and sequence-level clipping.

\section{EXPERIMENTS}
\label{sec:experiments}

\begin{table*}[t]
    \centering
    \caption{Performance Comparison on Navhard split in NAVSIM v2 using Closed-Loop Metrics.}
    \label{tab:navhard}
    
    \setlength{\tabcolsep}{7.5pt}
    \renewcommand{\arraystretch}{1.2}
    
    \begin{tabular}{l|c|ccccccccc|c|c}
        \hline
        \textbf{Method} & \textbf{Stage} &
        \textbf{NC$\uparrow$} &
        \textbf{DAC$\uparrow$} &
        \textbf{DDC$\uparrow$} &
        \textbf{TLC$\uparrow$} &
        \textbf{EP$\uparrow$} &
        \textbf{TTC$\uparrow$} &
        \textbf{LK$\uparrow$} &
        \textbf{HC$\uparrow$} &
        \textbf{EC$\uparrow$} &
        \textbf{S.$\uparrow$} &
        \textbf{EPDMS$\uparrow$} \\
        \hline
        
        \rowcolor{gray!15}
        \multicolumn{13}{c}{\textit{Traditional End-to-End Methods}} \\
        
        \multirow{2}{*}{LTF}
        & S1 & 96.2 & 79.6 & 99.1 & 99.6 & 84.1 & 95.1 & 94.2 & 97.6 & 79.1 & -- & \multirow{2}{*}{25.1} \\
        & S2 & 77.8 & 70.2 & 84.3 & 98.1 & 85.1 & 85.1 & 45.4 & 95.7 & 76.0 & -- & \\
        \cline{1-13}
        
        \multirow{2}{*}{GuideFlow}
        & S1 & 96.6 & 80.5 & 96.3 & 99.3 & 82.3 & 94.9 & 91.5 & 97.7 & 67.8 & -- & \multirow{2}{*}{\underline{27.1}} \\
        & S2 & 87.3 & 76.7 & 88.8 & 99.2 & 84.3 & 85.1 & 49.7 & 93.1 & 44.5 & -- & \\
        \hline
        
        \rowcolor{gray!15}
        \multicolumn{13}{c}{\textit{VLA-based Methods}} \\
        
        \multirow{2}{*}{DriveVLA-W0}
        & S1 & 96.8 & 83.3 & 99.0 & 99.6 & 84.6 & 95.3 & 96.4 & 97.6 & 78.2 & -- & \multirow{2}{*}{24.4} \\
        & S2 & 76.8 & 64.3 & 79.9 & 98.3 & 89.2 & 75.0 & 46.8 & 95.8 & 53.1 & -- & \\
        \cline{1-13}
        
        \multirow{2}{*}{ReCogDrive}
        & S1 & 96.4 & 78.9 & 98.7 & 99.8 & 82.6 & 95.6 & 94.4 & 97.6 & 74.2 & 67.7 & \multirow{2}{*}{25.7} \\
        & S2 & 80.2 & 65.0 & 82.4 & 98.7 & 85.2 & 76.9 & 43.8 & 96.6 & 71.8 & \underline{37.6} & \\
        \cline{1-13}
        
        \multirow{2}{*}{SGDrive}
        & S1 & 95.8 & 87.6 & 97.8 & 99.8 & 84.4 & 94.7 & 92.9 & 97.8 & 28.9 & \underline{71.1} & \multirow{2}{*}{25.5} \\
        & S2 & 79.4 & 65.4 & 79.1 & 98.9 & 88.9 & 75.3 & 42.7 & 96.4 & 29.6 & 35.2 & \\
        \hline
        
        \multirow{2}{*}{\textbf{CAR-VLA (Ours)}}
        & S1 & 96.0 & 92.2 & 98.8 & 99.6 & 86.5 & 94.9 & 94.7 & 97.6 & 73.8 & \textbf{79.3} & \multirow{2}{*}{\textbf{35.0}} \\
        & S2 & 79.0 & 77.9 & 85.3 & 97.7 & 91.8 & 74.7 & 52.2 & 95.0 & 57.8 & \textbf{43.5} & \\
        \hline
    \end{tabular}
    
    \par\vspace{2pt}
    \noindent
    \begin{minipage}{0.95\linewidth}
        \footnotesize
        \raggedright
        S. denotes the per-stage EPDM score. ${\uparrow}$ indicates higher is better. The best and second best results are \textbf{bold} and \underline{underlined}, respectively.
    \end{minipage}
\end{table*}

We evaluate CAR-VLA on NAVSIM v1, NAVSIM v2, and
the two-stage Navhard protocol, comparing its planning
performance with conventional end-to-end planners
and VLA-based methods.
We report both SFT and RL results to assess the
contribution of policy optimization.
We further analyze the generation length and inference
latency of the three reasoning modes in
Sec.~\ref{sec:exp_analysis}, and examine the effects
of reward components and Stage I pretraining in
Sec.~\ref{sec:exp_ablation}.

\subsection{Experimental Setup}
\label{sec:exp_setup}

\subsubsection{Training Data and Evaluation Protocols}

Our scene assessment and mode-adaptive driving datasets
are constructed from the same 103k scenes in the
\textit{navtrain} split%
~\cite{caesar2021nuplan,dauner2024navsim},
as described in Sec.~\ref{sec:data_curation}.
Stage I additionally uses the general driving data
provided by ReCogDrive~\cite{li2025recogdrive}.
We evaluate the SFT and RL policies on NAVSIM v1
and v2, following their respective metric suites.
For Navhard, we report the component metrics of both
evaluation stages and the overall EPDMS.
All comparisons are conducted within the corresponding
evaluation protocol.

\subsubsection{Metrics}

For NAVSIM v1, we report NC, DAC, EP, TTC, Comf,
and the aggregate PDMS.
With subscores normalized to $[0,1]$, the per-scene
PDMS is
\begin{equation}
\mathrm{PDMS}
=
\mathrm{NC}\cdot\mathrm{DAC}\cdot
\frac{5\mathrm{EP}+5\mathrm{TTC}+2\mathrm{Comf}}{12}.
\label{eq:pdms}
\end{equation}

NAVSIM v2 and Navhard use EPDMS, which combines
multiplicative penalties with weighted performance
terms:
\begin{equation}
\mathrm{EPDMS}
=
\left(\prod_{m\in\mathcal{M}}\tilde{s}_m\right)
\frac{\sum_{m\in\mathcal{W}}w_m\tilde{s}_m}
     {\sum_{m\in\mathcal{W}}w_m},
\label{eq:epdms}
\end{equation}
where
\[
\begin{aligned}
\mathcal{M}
&=\{\mathrm{NC},\mathrm{DAC},\mathrm{DDC},\mathrm{TLC}\},\\
\mathcal{W}
&=\{\mathrm{EP},\mathrm{TTC},\mathrm{LK},\mathrm{HC},
     \mathrm{EC}\}.
\end{aligned}
\]
The weights are $5$ for EP and TTC, and $2$ for
LK, HC, and EC.
Following the human-reference filtering protocol,
we use
\begin{equation}
\tilde{s}_m =
\begin{cases}
1,   & s_m^{\mathrm{human}}=0,\\
s_m, & \text{otherwise},
\end{cases}
\label{eq:metric_filter}
\end{equation}
where $s_m$ and $s_m^{\mathrm{human}}$ denote the
agent and human-reference subscores, respectively.
\begin{figure}[t]
    \centering
    \includegraphics[width=0.99\linewidth]{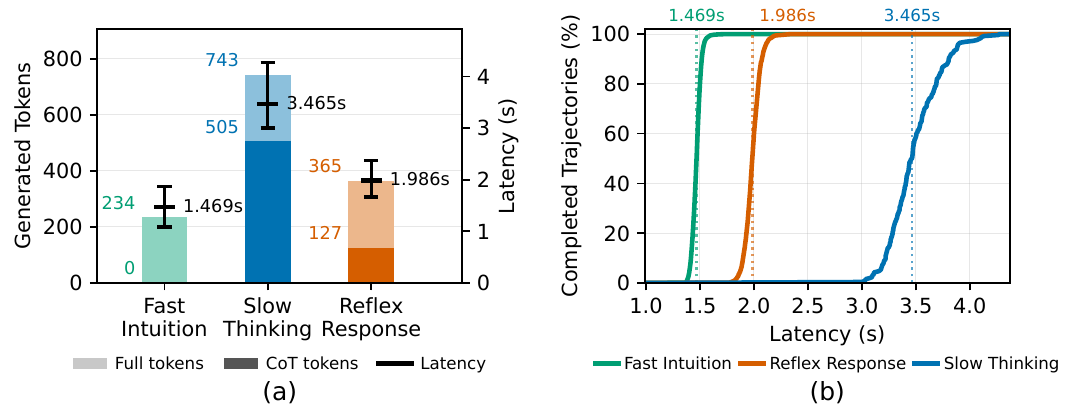}
    \caption{\textbf{Inference efficiency of CAR-VLA.}
(a) Cumulative percentage of \texttt{navtest} samples completing trajectory generation within a given latency.
(b) Total and chain-of-thought (CoT) token counts and inference latency across the three reasoning modes.}
    \label{fig:efficient}
\end{figure}

\begin{figure*}[t]
    \centering
    \includegraphics[width=0.99\linewidth]{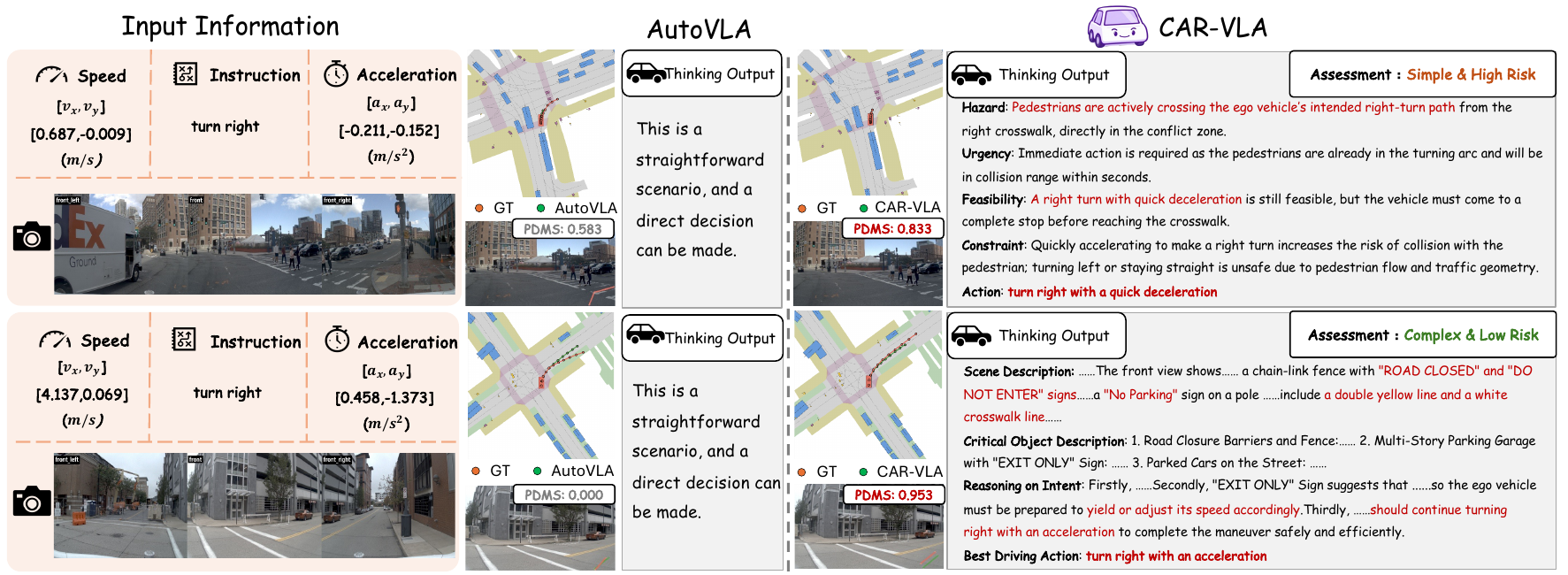}
    \caption{\textbf{Qualitative comparison of AutoVLA and CAR-VLA.}
CAR-VLA adapts its reasoning to simple, high-risk (top) and complex, low-risk (bottom) scenes through \emph{Reflex Response} and \emph{Slow Thinking}, respectively, achieving higher PDMS in both cases and correctly identifying the risk factors.}
    \label{fig:vis-navsim}
\end{figure*}

\begin{figure}[!t]
    \centering
    \includegraphics[width=0.90\linewidth]{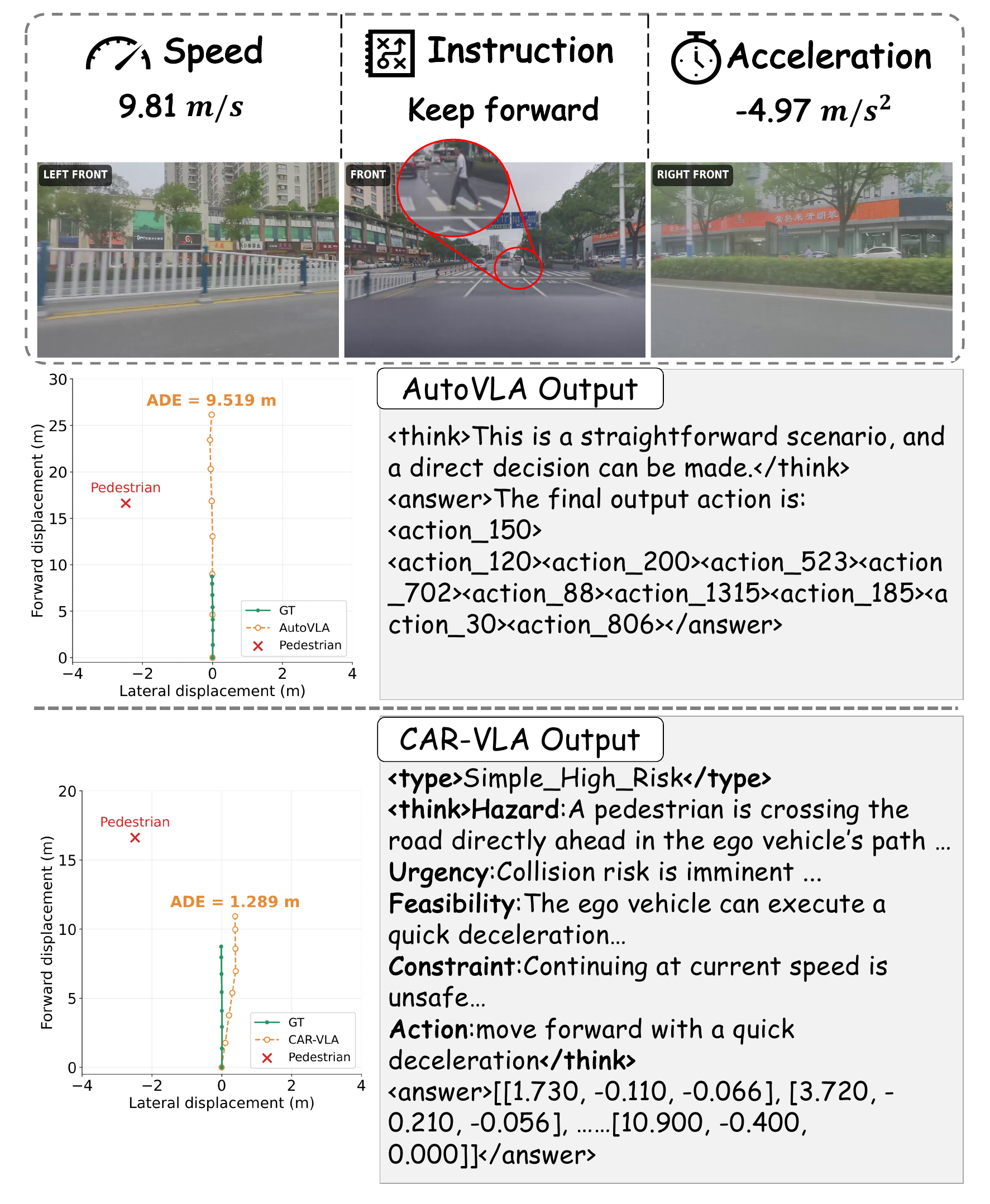}
    \caption{\textbf{Qualitative comparison in a pedestrian-crossing scenario.}
CAR-VLA recognizes the high risk despite low scene complexity and invokes \emph{Reflex Response}, producing a trajectory closer to the ground truth than AutoVLA (ADE: 1.289\,m vs.\ 9.519\,m).}
    \label{fig:vis-inhouse}
\end{figure}

\subsubsection{Implementation Details}
CAR-VLA is initialized from Qwen3-VL-4B~\cite{bai2025qwen3vl}.
For annotation, the Qwen3-VL-235B-A22B teacher receives four historical frames at 2\,Hz from each front-left, front, and front-right camera (12 images per scene). SFT Stage II training uses one frame per view (3 images).
Following AutoVLA~\cite{zhou2025autovla}, fixed rules map ground-truth future ego speed and acceleration to action hints used only for offline driving-reasoning annotation.
CAR-VLA predicts ten future ego poses at 0.5\,s intervals over 5\,s.

Both SFT stages use eight NVIDIA H200 GPUs.
Stage~I freezes the vision encoder and projector with a learning rate of $10^{-5}$; Stage~II trains all parameters at $2\times10^{-5}$.
Reasoning-augmented RL uses Qwen3-Embedding-0.6B~\cite{qwen3embedding} for sentence embedding, a learning rate of $2\times10^{-6}$, $\lambda_{\mathrm{d}}$=0.7, $\lambda_{\mathrm{g}}$=0.2, $\lambda_{\mathrm{r}}$=0.1, rollout/global batch sizes of 256/128, and 10k hard scenes selected for high rollout-reward variance or low mean reward from Navtrain following ELF-VLA~\cite{luo2026elfvla}.
We evaluate after Stage~II SFT and RL.


\subsection{Main Results}
\label{sec:exp_main}

\subsubsection{NAVSIM v1 and v2 Navtest}

Table~\ref{tab:navsim_main} compares CAR-VLA with end-to-end and VLA-based baselines. CAR-VLA-RL achieves the highest NAVSIM v2 EPDMS among the compared methods at 90.3, surpassing IRR-Drive by 1.3 points, while obtaining a competitive PDMS of 91.1 on NAVSIM v1. Compared with CAR-VLA-SFT, reinforcement learning improves the aggregate scores by 3.0 and 1.8 points on v1 and v2, respectively, alongside gains in ego progress and drivable-area compliance.

\subsubsection{NAVSIM v2 Navhard}

Table~\ref{tab:navhard} presents the Navhard results. CAR-VLA achieves an overall EPDMS of 35.0, exceeding the strongest compared baseline, GuideFlow, by 7.9 points. It also achieves the highest reported scores in both Stage~1 and Stage~2, at 79.3 and 43.5, respectively. Together with the leading DAC and EP scores in both stages, these results demonstrate CAR-VLA's competitive planning performance on challenging scenarios, particularly in maintaining progress and drivable-area compliance.

\subsection{Adaptive-Reasoning Analysis}
\label{sec:exp_analysis}
Fig.~\ref{fig:efficient}(a) compares generation length and latency across the three reasoning modes. Fast Intuition generates 234 total tokens without an explicit reasoning trace and has a reported latency of 1.469~s. Slow Thinking generates 743 total tokens, including 505 reasoning tokens, with a latency of 3.465~s. Reflex Response occupies an intermediate regime, producing 365 total tokens with 127 reasoning tokens and a latency of 1.986~s.

Compared with Slow Thinking, Reflex Response reduces reasoning tokens by 74.9\% and reported latency by 42.7\%, while retaining explicit reasoning. Fast Intuition further reduces latency by omitting the reasoning trace. The latency distributions in Fig.~\ref{fig:efficient}(b) show the same ordering, with Fast Intuition and Reflex Response outputs completed earlier than Slow Thinking outputs. These observations support the intended differentiation in computational cost: detailed deliberation incurs the largest generation overhead, while hazard-focused reasoning provides a shorter intermediate response. The figure characterizes generation efficiency and the planning comparisons are reported separately in Sec.~\ref{sec:exp_main}.

\subsection{Ablation Studies}
\label{sec:exp_ablation}

\subsubsection{Effect of Reward Components}
\begin{table}[t]
    \centering
    \caption{Ablation study of RL reward components. Values in parentheses denote EPDMS changes relative to the baseline.}
    \label{tab:rl_reward_ablation}
    \setlength{\tabcolsep}{2.0pt}
    \resizebox{\columnwidth}{!}{
    \begin{tabular}{l|ccccccccc|c}
        \toprule
        \textbf{Reward}
        & \textbf{NC}
        & \textbf{DAC}
        & \textbf{DDC}
        & \textbf{TLC}
        & \textbf{EP}
        & \textbf{TTC}
        & \textbf{LK}
        & \textbf{HC}
        & \textbf{EC}
        & \textbf{EPDMS} \\
        \midrule
        SFT(baseline)
        & 98.6 & 96.6 & 99.5 & 99.9 & 86.3 & 98.0 & 97.4 & 98.3 & 86.0
        & 88.5 \\
        \midrule
        $R_{\mathrm{d.}} + R_{\mathrm{f.}}$
        & 97.1 & 95.5 & 98.9 & 99.8 & 93.5 & 96.7 & 96.2 & 97.9 & 84.7
        & 87.2 {\scriptsize (-1.3)} \\

        $+ R_{\mathrm{g.}}$
        & 98.4 & 97.8 & 99.3 & 99.8 & 88.2 & 97.7 & 96.9 & 98.2 & 86.4
        & 89.7 {\scriptsize (+1.2)} \\

        $+ R_{\mathrm{r.}}$
        & 98.5 & 98.1 & 99.4 & 99.9 & 89.0 & 97.9 & 97.4 & 98.3 & 85.8
        & \textbf{90.3 {\scriptsize (+1.8)}}  \\
        \bottomrule
    \end{tabular}
    }
\end{table}

Table~\ref{tab:rl_reward_ablation} evaluates the progressive addition of reward components on NAVSIM v2. Optimizing the driving and format rewards alone yields 87.2 EPDMS, 1.3 points below the SFT baseline. Although EP increases from 86.3 to 93.5, NC, DAC, and TTC decrease from 98.6, 96.6, and 98.0 to 97.1, 95.5, and 96.7, respectively. This configuration therefore improves progress at the expense of other driving criteria and does not improve the aggregate score.

Adding the geometry reward raises EPDMS to 89.7, a gain of 2.5 points over the basic reward configuration. NC, DAC, and TTC recover to 98.4, 97.8, and 97.7, while EP decreases to 88.2. This pattern is consistent with the terminal-geometry objective providing an additional constraint on trajectory optimization. Adding the final reasoning reward term, denoted by $R_{\mathrm{r.}}$ in the table, further increases EPDMS to 90.3, with improvements in EP, DAC, TTC, and LK over the geometry-augmented configuration. The complete reward achieves a 1.8-point gain over SFT. These cumulative comparisons show the incremental benefit of the auxiliary terms within the reported reward sequence.

\subsubsection{Effect of Stage I SFT}
\begin{table}[t]
    \centering
    \caption{Ablation study on CAR-VLA Components.}
    \label{pretraining_ablation}
    \setlength{\tabcolsep}{2.0pt}
    \resizebox{\columnwidth}{!}{
    \begin{tabular}{l|ccccccccc|c}
        \toprule
        \textbf{Model}
        & \textbf{NC}
        & \textbf{DAC}
        & \textbf{DDC}
        & \textbf{TLC}
        & \textbf{EP}
        & \textbf{TTC}
        & \textbf{LK}
        & \textbf{HC}
        & \textbf{EC}
        & \textbf{EPDMS} \\
        \midrule
        SFT w/o Stage I
        & 98.4 & 96.4 & 99.4 & 99.9 & 86.2 & 97.8 & 97.4 & 98.3 & 85.9
        & 88.1 \\

        \quad + RL
        & 98.4 & 97.8 & 99.4 & 99.8 & 88.2 & 97.8 & 96.9 & 98.3 & 86.4
        & 89.9 \\

        \midrule
        SFT w/ Stage I
        & 98.6 & 96.6 & 99.5 & 99.9 & 86.3 & 98.0 & 97.4 & 98.3 & 86.0
        & 88.5 \\

        \quad + RL
        & 98.5 & 98.1 & 99.4 & 99.9 & 89.0 & 97.9 & 97.4 & 98.3 & 85.8
        & \textbf{90.3} \\
        \bottomrule
    \end{tabular}
    }
\end{table}

Table~\ref{pretraining_ablation} examines stage I SFT training before the subsequent stage II SFT and RL. Stage I SFT improves EPDMS from 88.1 to 88.5 and the post-RL score from 89.9 to 90.3, giving a consistent 0.4-point improvement in both settings. After RL, the model also achieves higher EP, DAC, and LK than its counterpart without stage I training, although the gains do not extend to every component metric.

Reinforcement learning adds 1.8 EPDMS points with or without Stage I SFT. Thus, the advantage is retained after policy optimization, while RL provides a further improvement in either setting. The strongest configuration combines 2-stage SFT with RL and reaches 90.3 EPDMS.

\subsection{Qualitative Analysis}


\subsubsection{Comparison on Navtest}

Fig.~\ref{fig:vis-navsim} compares AutoVLA
and CAR-VLA on two \textit{navtest} scenes with the
same right-turn instruction but different reasoning
requirements.
AutoVLA treats both scenes as straightforward and
opts for direct planning, whereas CAR-VLA selects
different reasoning modes based on its complexity--risk
assessment.

\noindent\textbf{Simple high-risk scene (top).}
CAR-VLA classifies the scene as simple--high-risk
and invokes \textit{Reflex Response}.
Its reasoning identifies pedestrians crossing the
intended right-turn path as the critical hazard,
emphasizes the need to stop before the crosswalk,
and recommends a right turn with rapid deceleration.
This contrasts with AutoVLA's generic direct-decision
response, which provides no explicit account of the
pedestrian conflict.
The resulting CAR-VLA trajectory achieves a PDMS
of $0.833$, compared with $0.583$ for AutoVLA.

\noindent\textbf{Complex low-risk scene (bottom).}
CAR-VLA predicts complex--low-risk and adopts
\textit{Slow Thinking}.
Its reasoning considers road-closure signs, lane
markings, and a parking-garage exit, relating these
environmental cues to the intended maneuver before
recommending a right turn with acceleration.
CAR-VLA achieves a PDMS of $0.953$, whereas
AutoVLA scores $0.000$.

These cases illustrate the distinction
between interaction urgency and environmental
interpretation demands: CAR-VLA focuses on the
immediate pedestrian conflict in the former and
performs broader structural and regulatory analysis
in the latter, with higher planning scores in
both examples.

\subsubsection{In-House High-Risk Scenario}

Fig.~\ref{fig:vis-inhouse} presents a
pedestrian-crossing case from our in-house data (involving an AEB scenario).
AutoVLA treats the scene as straightforward and
predicts substantially greater forward displacement
than the ground truth.
In contrast, CAR-VLA classifies it as simple--high-risk
and invokes \textit{Reflex Response}, identifying
the pedestrian conflict and recommending rapid
deceleration.
Its trajectory more closely matches the ground truth,
achieving an ADE of $1.289\,\mathrm{m}$ compared
with $9.519\,\mathrm{m}$ for AutoVLA. In terms of the outcome, CAR-VLA's timely deceleration ultimately avoids a collision with the pedestrian.


\section{CONCLUSION}
\label{sec:conclusion}

We presented CAR-VLA, a unified driving VLA that adapts
reasoning depth and focus to static complexity and
dynamic risk. It maps four scene categories to
Fast Intuition, Slow Thinking, and Reflex Response,
with a dedicated hazard-focused reasoning process
for high-risk situations.
Complexity--risk-guided data curation and two-stage
supervised fine-tuning connect scene assessment with
mode-specific reasoning and trajectory generation,
while reasoning-augmented reinforcement learning
further optimizes driving performance and reasoning
alignment.
Experiments on NAVSIM v1, NAVSIM v2, and Navhard
demonstrate competitive planning performance, while
qualitative cases illustrate responses to identified
hazards.
Our findings highlight the value of adapting not
only whether to reason, but also how to reason
according to the demands of the driving scene.

\section*{Acknowledgments}
The author used ChatGPT (OpenAI) to polish the language and provided suggestions on chart layout, cartoon icon design, and presentation methods. The research methodology, experimental design, analysis, and conclusions were independently developed and verified by the authors.

\bibliographystyle{IEEEtran}
\bibliography{references}

\end{document}